\documentclass[10pt,twocolumn,letterpaper]{article}

\usepackage{iccv}
\usepackage{times}
\usepackage{epsfig}
\usepackage{graphicx}
\usepackage{amsmath}
\usepackage{amssymb}
\usepackage{authblk}

\usepackage{graphicx}
\usepackage{booktabs}
\usepackage{multirow}
\usepackage[ruled,linesnumbered]{algorithm2e}
\newtheorem{definition}{definition}
\makeatletter
\newcommand{\printfnsymbol}[1]{%
	\textsuperscript{\@fnsymbol{#1}}%
}
\usepackage[breaklinks=true,bookmarks=false]{hyperref}
\usepackage{url}
\iccvfinalcopy 

\def\iccvPaperID{****} 

\ificcvfinal\fi

\begin{document}

\title{T-SVDNet: Exploring High-Order Prototypical Correlations for Multi-Source Domain Adaptation}
\author{Ruihuang Li$^1$\thanks{Work partly done during an internship at Noah’s Ark Lab}, Xu Jia$^2$\thanks{Corresponding Author}, Jianzhong He$^3$, Shuaijun Chen$^4$, Qinghua Hu$^1$\printfnsymbol{2}   \\ \vspace{-0.8em}
	$^1$Tianjin University, $^2$Dalian University of Technology, $^3$Huawei Technologies, $^4$Noah’s Ark Lab, Huawei Technologies \\ 
	{\tt\small \{liruihuang, huqinghua\}@tju.edu.cn}, {\tt\small xjia@dlut.edu.cn}, {\tt\small chensj1110@163.com}, {\tt\small jianzhong.he@huawei.com} }

\maketitle
\ificcvfinal\thispagestyle{empty}\fi

\begin{abstract}
	Most existing domain adaptation methods focus on adaptation from only one source domain, however, in practice there are a number of relevant sources that could be leveraged to help improve performance on target domain. We propose a novel approach named T-SVDNet to address the task of Multi-source Domain Adaptation (MDA), which is featured by incorporating Tensor Singular Value Decomposition (T-SVD) into a neural network's training pipeline. Overall, high-order correlations among multiple domains and categories are fully explored so as to better bridge the domain gap. Specifically, we impose Tensor-Low-Rank (TLR) constraint on a tensor obtained by stacking up a group of prototypical similarity matrices, aiming at capturing consistent data structure across different domains. Furthermore, to avoid negative transfer brought by noisy source data, we propose a novel uncertainty-aware weighting strategy to adaptively assign weights to different source domains and samples based on the result of uncertainty estimation. Extensive experiments conducted on public benchmarks demonstrate the superiority of our model in addressing the task of MDA compared to state-of-the-art methods. Code is available at {\small  \url{https://github.com/lslrh/T-SVDNet}}.
\end{abstract}
\section{Introduction}
Deep learning methods have shown superior performance with huge amounts of training data as rocket fuel. However, directly transferring knowledge learned on a certain visual domain to other domains with different distributions would degrade the performance significantly due to the existence of domain shift \cite{yosinski2014transferable}. To handle this problem, the prominent approaches such as transfer learning and unsupervised domain adaptation (UDA) endeavor to extract domain-invariant features. Discrepancy-based methods reduce the domain gap by minimizing the discrepancy between source and target distributions, such as Maximum Mean Discrepancy (MMD) \cite{long2015learning}, correlation alignment \cite{sun2017correlation}, and contrastive domain discrepancy \cite{kang2019contrastive}. Adversarial methods attempt to align source and target domains through adversarial training \cite{saito2018maximum,tzeng2017adversarial} or GAN-based loss \cite{hoffman2018cycada,zhu2017unpaired}. These methods only focus on domain adaptation with only single source. However, in many practical application scenarios, there are a number of relevant sources collected in different ways available, which could be used to help improve performance on target domain. \par
Naively combining various sources into one is not an effective way to fully exploit abundant information within multiple sources, and might even perform worse than single-source methods, because domain gap among multiple sources causes confusion in the learning process \cite{zhao2020}. Some Multi-Source Domain Adaptation (MDA) approaches~\cite{xu2018deep,li2018extracting,zhao2020multi,peng2019moment,guo2018multi} focus on aligning multiple source domains and a target domain by projecting them into a domain-invariant feature space. This is done by either explicitly minimizing the discrepancy of different domains \cite{guo2018multi,hoffman2018algorithms,peng2019moment} or learning an adversarial discriminator to align distributions of different domains \cite{xu2018deep,zhao2018adversarial,li2018extracting}. However, eliminating distribution discrepancy of data has the risk of sacrificing discrimination ability. Moreover, these methods only achieve pair-wise matching, neglecting underlying high-order relations among all domains. Another widely used way in MDA is distribution-weighted combining rule~\cite{hoffman2018algorithms,zhao2018adversarial, li2018extracting}, which takes a weighted combination of pre-trained source classifiers as the classifier for target domain. In spite of reasonable performance on MDA task, they do not take into consideration intra-domain weightings among different training samples, so that underlying noisy source data may hurt the performance of learning in the target, which is referred to as ``negative transfer"~\cite{pan2009survey}.\par
To address the aforementioned limitations, we propose a novel method named T-SVDNet which incorporates tensor singular value decomposition into a neural network's training pipeline. In MDA tasks, although there is large domain gap between different domains, data belonging to the same category do share essential semantic information across domains. Therefore, we assume that data from different domains should follow a certain kind of category-wise structure. Based on this assumption, we explore high-order relationships among multiple domains and categories in order to enforce the alignment of source and target at the prototypical correlation level. Specifically, we impose Tensor-Low-Rank (TLR) constraint on a tensor which is obtained by stacking up a set of prototypical similarity matrices, so that the relationships between categories are enforced to be consistent across domains by pursuing the lowest-rank structure of tensor. Furthermore, to avoid negative transfer \cite{pan2009survey} caused by noisy training data, we propose a novel uncertainty-aware weighting strategy to guide the adaptation process. It could dynamically assign weights to different domains and training samples based on the result of uncertainty estimation. To train the whole framework with both classification loss and low-rank regularizer, we adopt an alternative optimization strategy, that is, optimizing network parameters with the low-rank tensor fixed and optimizing the low-rank tensor with network parameters unchanged. We conduct extensive evaluations on several public benchmark datasets, where a significant improvement over existing MDA methods has been achieved. Overall, the main contributions of this paper can be summarized as follows:\\
$\bullet$ We propose the T-SVDNet to explore high-order relationships among multiple domains and categories from the perspective of tensor, which facilitates both domain-invariance and category-discriminability.\\
$\bullet$ We devise a novel uncertainty-aware weighting strategy to balance different source domains and samples, so that clean data are fully exploited while negative transfer led by noisy data is avoided. \\
$\bullet$ We propose an alternative optimization method to train the deep model along with low-rank regularizer. Extensive evaluations on benchmark datasets demonstrate the superiority of our method. 
\begin{figure*}
	\centering
	\includegraphics[scale=0.28]{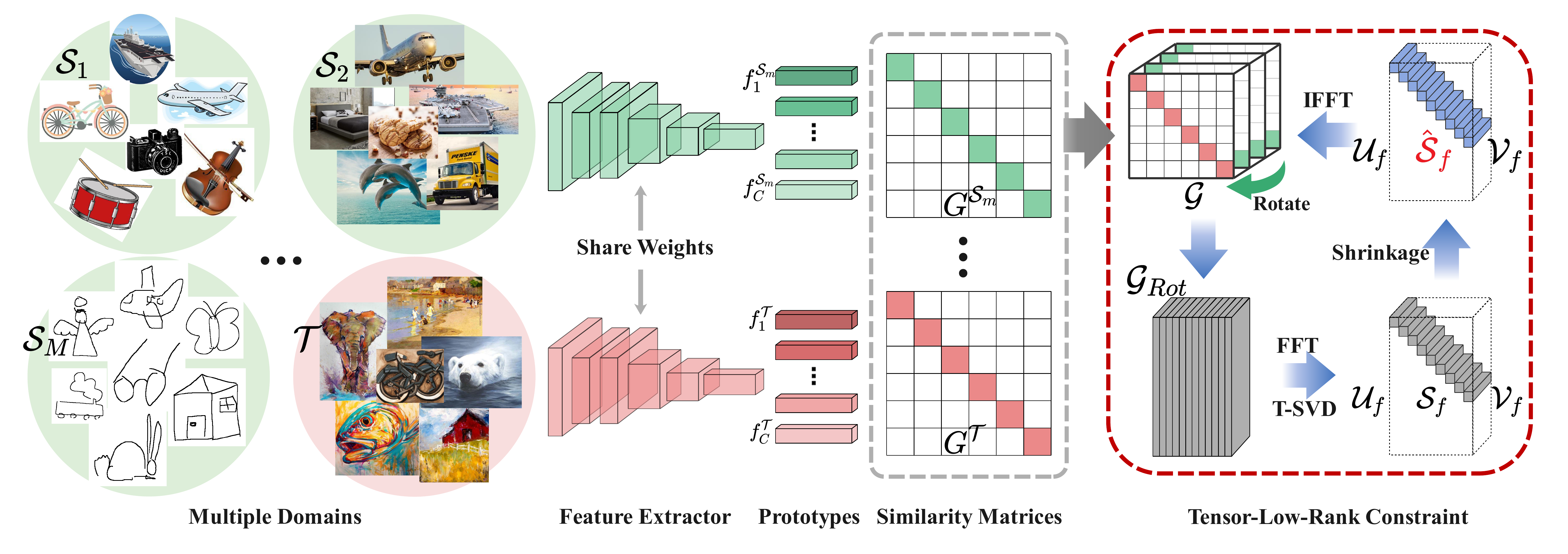}\\
	\caption{The framework of T-SVDNet. Given $M$ labeled source domains $\mathcal{S}_1, \cdots, \mathcal{S}_M$ and an unlabeled target domain $\mathcal{T}$, we first extract features for input images and compute prototype $f_c^{\mathcal{D}}$ for each category and each domain in an online fashion. Furthermore, the relations between pairwise prototypes are modeled by a group of similarity matrices ${G}^{\mathcal{S}_1},\cdots, {G}^{\mathcal{S}_M}, {G}^{\mathcal{T}}$. Then we stack these prototypical similarity matrices into a 3-order tensor $\mathcal{G} \in \mathbb{R}^{C\times C\times (M+1)}$ on which tensor-low-rank constraint is imposed in order to explore high-order relationships among different domains. Finally, together with low-rank regularizer, the model is effectively trained in an alternative optimization strategy. }
	\label{fig1}
	\vspace{-1em}
\end{figure*}


\section{Related Work}
\label{sec2}
\textbf{Single-source Domain Adaptation (SDA).} SDA aims to generalize a model learned from a labeled source domain to a related unlabeled domain with different data distribution. Existing SDA methods usually incorporate two terms: one term is task loss like cross-entropy loss which helps learn a model on the labeled source; the other adaptation term aims to align the distributions of source and target domains. These SDA methods can be roughly categorized into three groups according to the alignment strategies: (1) discrepancy-based methods aim to minimize the discrepancy which is explicitly measured on corresponding layers, including Maximum Mean Discrepancy (MMD) \cite{long2015learning}, correlation alignment \cite{sun2017correlation}, and contrastive domain discrepancy \cite{kang2019contrastive}; (2) Some adversarial-based methods align different data distributions by confusing a well-trained domain discriminator \cite{tzeng2017adversarial,tsai2018learning}. In addition, adversarial generative methods aggregate domains at pixel level by generating adapted fake data \cite{zhu2017unpaired}; (3) Reconstruction-based methods propose to reconstruct the target domain from latent representation by using the source task model \cite{ghifary2016deep}. \par
\textbf{Multi-source Domain Adaptation (MDA).} In practical applications, data may be collected from multiple related domains \cite{bhatt2016multi,sun2015survey}, which involve more abundant information but also bring the difficulty in handling the domain shift. Thus MDA methods become more and more popular. The earlier MDA methods mainly focus on weighted combination of source classifiers \cite{hoffman2018algorithms,li2018extracting,lee2019sliced,saito2018maximum} based on the assumption that target distribution can be approximated by the mixture of source distributions \cite{blitzer2007learning,ben-david2010a}. Hoffman \etal \cite{hoffman2018algorithms} cast distribution combination as a DC-programming and derived a tighter domain generalization bound. Besides classification losses, various domain assignment constraints are devised to reduce the domain gap. In addition to minimizing domain discrepancy between the target and each source domain, Li \etal \cite{li2018extracting} also took into consideration the relationships between pairwise source domains and proposed a tighter bound on the discrepancy among multiple sources. Many explicit measures of discrepancy have been used in MDA methods, such as MMD \cite{guo2018multi}, $\mathcal{L}_2$ distance \cite{rakshit2019unsupervised}, and moment distance \cite{peng2019moment}. Some approaches also focus on prototype-based alignment between different domains \cite{pan2019transferrable,xie2018learning,wang2020learning}. As for the adversarial MDA methods which aim to confuse the discriminator so that domain-invariant features are extracted, the optimized objective can be $\mathcal{H}$-divergence \cite{zhao2018adversarial}, Wasserstein distance \cite{zhao2020multi,li2018extracting}. \par
\textbf{Uncertainty Estimation.} 
Quantifying and measuring uncertainty is of great theoretical and practical significance~\cite{faber2005on,kiureghian2009aleatory}. In Bayesian modeling, there are two main categories of uncertainty~\cite{kendall2017what}: epistemic uncertainty and aleatoric uncertainty. The former is often referred to as model uncertainty, which captures uncertainty in the model parameters, while the latter accounts for noise inherent from the observations. There have been many methods proposed to estimate uncertainty in deep learning~\cite{blundell2015weight,gal2016dropout,cipolla2018multi}. Resorting to these techniques, the robustness and interpretability of many computer vision tasks are improved, such as object detection \cite{choi2019gaussian,kraus2019uncertainty} and face recognition \cite{chang2020data}.        
\section{Method}
In the MDA setting, there are $M$ labeled source domains ${\mathcal{S}}_1, {\mathcal{S}}_2, \cdots, {\mathcal{S}}_M$ and an unlabeled target domain ${\mathcal{T}}$. Each source domain ${\mathcal{S}}_m$ contains $N_m$ observations $\{({x}_{i}^{\mathcal{S}_m}, {y}_{i}^{\mathcal{S}_m})\}_{i=1}^{N_m} $, where $y_{i}$ is the desired label, while in the target domain ${\mathcal{T}}$, the label $y$ is not available. Most existing MDA models can be formulated as the following mapping function:
\begin{align}
	M_{mda}: X^{{\mathcal{S}}_1}\cup \cdots X^{{\mathcal{S}}_M} \cup X^{\mathcal{T}}\rightarrow Y^{{\mathcal{S}}_1}\cup\cdots Y^{{\mathcal{S}}_M},
\end{align}
where $M_{mda}$ is trained on both labeled samples $(X^{\mathcal{S}}, Y^{\mathcal{S}})$ in the source domain and unlabeled samples $X^{\mathcal{T}}$ in the target domain. \par
In this section, we propose the T-SVDNet which fully explores high-order relationships among all domains by exploiting the tensor obtained by stacking up a set of prototypical similarity matrices (see Fig.~\ref{fig1}). In addition, we propose a novel uncertainty-aware weighting strategy to achieve both inter- and intra-domain weightings so that negative transfer is reduced (see Fig.~\ref{fig2}). This section is organized as follows: we first construct prototypical similarity matrix in Sec.~\ref{sec3.1}. Then we propose the tensor-low-rank constraint and uncertainty-aware weighting strategy in Sec.~\ref{sec3.2} and Sec.~\ref{sec3.3}, respectively. Finally, we formulate the total objective function in Sec.~\ref{sec3.4} and propose a novel alternative optimization method in Sec.~\ref{sec3.5}.
\subsection{Prototypical similarity matrix}
\label{sec3.1}
In the proposed T-SVDNet, we first map input image into latent space through a feature extractor denoted by $f(\cdot)$, then we update the centroid of each category (prototype) based on the feature embeddings of a mini-batch \cite{pan2019transferrable,xie2018learning,wang2020learning}. For domain ${\mathcal{D}}\in \{\mathcal{S}_1, \cdots, \mathcal{S}_M, \mathcal{T}\}$, the prototype of the $c$-th category denoted by $f_c^{\mathcal{D}}$ is computed by: 
\begin{align}
	f_{c}^{{\mathcal{D}}}=\frac{1}{\left | {\Lambda}_{c}^{{\mathcal{D}}} \right |}\sum_{(x_i, y_i)\in {\Lambda}_{c}^{{\mathcal{D}}}} f(x_{i}),
\end{align} 
where ${\Lambda}_{c}^{{\mathcal{D}}}$ is the set of training samples belonging to the $c$-th category in domain ${\mathcal{D}}$, \ie, ${\Lambda}_{c}^{{\mathcal{D}}} = \{(x_i, y_i)\in {\mathcal{D}}|y_{i} = c \}$. It is noteworthy that for unlabeled target domain, we first assign pseudo labels $\hat{y}_i$ to samples with high classification confidence. Specifically, we first map each image in target domain $x_i^{\mathcal{T}}$ into a classification probability vector $p_i$, then we set a threshold $\tau$ for selecting confident predictions as pseudo labels $\hat{y}_i^{(k)}$, \ie,
\begin{align}
	\hat{y}_i^{(k)*}=\left\{\begin{matrix}
		1, & if\ k=\underset{c}{\rm argmax}\;p_i^{(c)}\ {\rm and}\ p_i^{(k)}>\tau\\ 
		0, & {\rm otherwise}
	\end{matrix}\right..
\end{align}
\par
In order to reduce the randomness in sampling of each mini-batch and stabilize the training process, the category prototypes are updated according to exponential moving average (EMA) method:
\begin{align}
	f_{c}^{\mathcal{D}}|_I:=\alpha f_{c}^{\mathcal{D}}|_I+(1-\alpha)f_{c}^{\mathcal{D}}|_{I-1},
\end{align}
where $\alpha$ is the exponential decay rate and $I$ denotes current iteration. \par
Then we employ Gaussian kernel to model inter-class relationships and construct a series of prototypical similarity matrices ${G}^{\mathcal{S}_1},\cdots, {G}^{\mathcal{S}_M}, {G}^{\mathcal{T}}$: 
\begin{align}
	G_{c_i,c_j}^{\mathcal{D}}={\mathcal{K}}(f_{c_i}^{\mathcal{D}}, f_{c_j}^{\mathcal{D}}) = exp(-\frac{\left \|f_{c_i}^{\mathcal{D}}-f_{c_j}^{\mathcal{D}} \right \|_{2}^{2}}{2{\gamma}^2}),
\end{align}  
where $f_{c_i}^{\mathcal{D}}$ and $f_{c_j}^{\mathcal{D}}$ are a pair of category centroids from domain $\mathcal{D}$, and $\gamma$ is the deviation parameter which is set as $0.05$ in experiments.\par 

\subsection{Tensor-low-rank constraint via T-SVD}
\label{sec3.2}
Unlike conventional methods only considering pairwise matching, we achieve high-order alignment of all domains at the prototypical correlation level. Specifically, we stack prototypical similarity matrices into a 3-order tensor ${\mathcal{G}} \in {\mathbb{R}}^{C\times C \times (M+1)}$ along the third dimension, where $C$ and $M$ denote the number of classes and domains, respectively. Then we impose the Tensor-Low-Rank (TLR) constraint on the assembled tensor in order to explore high-order correlations among domains and enforce the relationships between categories to be consistent across domains. Here we first give definitions of T-SVD and tensor rank as follows:\par
\begin{definition}
	(T-SVD) Given tensor ${\mathcal{G}} \in {\mathbb{R}}^{n_1\times n_2 \times n_3}$, the tensor singular value decomposition of ${\mathcal{G}}$ is defined as a finite sum of outer product of matrices \cite{martin2013an}:
	\begin{align}
		\mathcal{G}=\sum_{i=1}^{min(n_1,n_2)}\mathcal{U}(:,i,:)*\mathcal{S}(i,i,:)*\mathcal{V}(:,i,:)^T,
	\end{align}
	where $\mathcal{U}$ and $\mathcal{V}$ are orthogonal tensors with size $n_1\times n_1\times n_3$ and $n_2\times n_2\times n_3$, respectively. $\mathcal{S}$ is a tensor with the size $n_1\times n_2\times n_3$, each frontal slice of which is a diagonal matrix. $*$ denotes tensor product (T-product). 
\end{definition}\par
T-SVD also can be computed more efficiently in the Fourier domain. Specifically, it can be replaced by conducting fast Fourier transformation (FFT) along the third dimension of $\mathcal{G}$ to get $\mathcal{G}_f$, and performing matrix SVDs on each frontal slice of $\mathcal{G}_f$:
\begin{align}
	\label{eq8}
	\mathcal{G}_f^{(k)} = \mathcal{U}_f^{(k)}\cdot \mathcal{S}_f^{(k)} \cdot \mathcal{V}_f^{(k)T},k=1,\cdots,n_3
\end{align}
where $\cdot$ means matrix product. We use $\mathcal{G}_f^{(k)}$ to denote the $k$-th frontal slice of $\mathcal{G}$, \ie, $\mathcal{G}_f^{(k)} = \mathcal{G}(:,:,k)$. The result of T-SVD is finally obtained by taking the inverse FFT on $\mathcal{U}_f, \mathcal{S}_f, \mathcal{V}_f$ along the third dimension (see Alg. \ref{alg1}).

\begin{algorithm}[!t]
	\SetKwInOut{Input}{Input}\SetKwInOut{Output}{Output}
	\caption{T-SVD}\label{alg1}
	\Input{${\mathcal{G}} \in {\mathbb{R}}^{n_1\times n_2 \times n_3}$\;}
	\Output{$\mathcal{U}$, $\mathcal{S}$, $\mathcal{V}$ \;}
	${\mathcal{G}_f} = \text{FFT}(\mathcal{G}, 3)$\;
	\For{$k=1:n_3$}{$[\mathcal{U}_f^{(k)},\mathcal{S}_f^{(k)}, \mathcal{V}_f^{(k)}] = \text{SVD}(\mathcal{G}_{f}^{(k)})$\;}
	$\mathcal{U}, \mathcal{S}, \mathcal{V} = \text{IFFT}(\mathcal{U}_f, 3), \text{IFFT}(\mathcal{S}_f,3), \text{IFFT}(\mathcal{V}_f,3)$ \;
\end{algorithm}

\begin{definition}
	(Tensor~rank)~\cite{kilmer2013third,lu2019tensor} The rank of $\mathcal{G}\in {\mathbb{R}}^{n_1\times n_2 \times n_3}$ is a vector ${\bf{p}}\in\mathbb{R}^{n_3\times1}$ with the $k$-th element equal to the rank of the $k$-th frontal slice of $\mathcal{G}_f^{(k)}$. 
\end{definition}

However, we need an adequate convex relaxation to $\ell_1$ norm of tensor rank in optimization process. To this end we formulate it as tensor nuclear norm, which is defined as the sum of the singular values of all frontal slices $\mathcal{S}_f^{(k)}$:
\begin{align}
	\left \| \mathcal{G} \right \|_{TNN} = \sum_{k=1}^{n_3}\sum_{i=1}^{min(n_1,n_2)}|\mathcal{S}_f^{(k)}(i,i)|. 
\end{align}
\textbf{Tensor rotation.} In view of each frontal slice $\mathcal{G}^{(k)}$ of tensor $\mathcal{G}$ only contains information from single domain, we rotate it horizontally (or vertically) to obtain ${\mathcal{G}}_{Rot}$ (see Fig. \ref{fig1}). In this way, each frontal slice ${\mathcal{G}}_{Rot}^{(k)}$ will involve information from different domains. In the end, imposing tensor-low-rank constraint on the rotated tensor ${\mathcal{G}}_{Rot}$ benefits the exploration of high-order relationships among different domains. The inter-category data structure is enforced to be consistent across domains by pursuing the lowest rank of each frontal slice ${\mathcal{G}}_f^{(k)}$ in Fourier domain.
\begin{figure}
	\centering
	\includegraphics[scale=0.6]{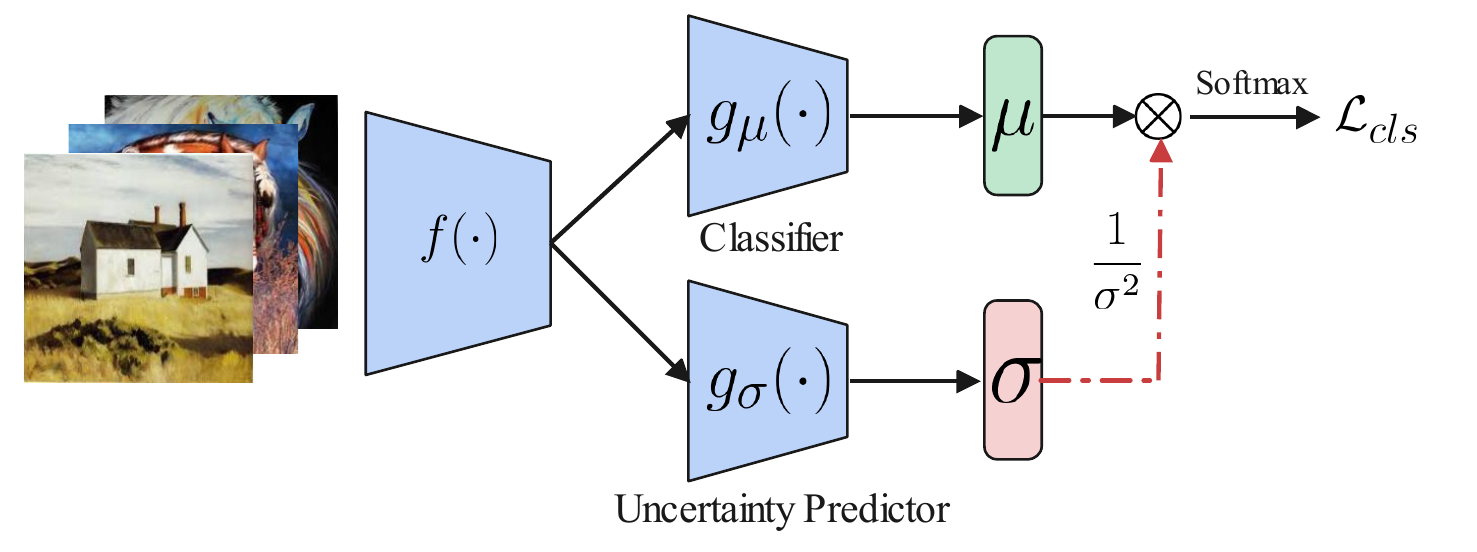}\\
	\caption{Uncertainty-aware weighting strategy. Each sample is modeled as a Gaussian distribution parameterized by mean $\mu$ and variance $\sigma$. The classification loss is weighted by estimated data uncertainty $\sigma$.}
	\label{fig2}
\vspace{-1em}
\end{figure}
\subsection{Uncertainty-aware weighting strategy}
\label{sec3.3}
Instead of equally treating each source domain and sample, we propose a novel uncertainty-aware weighting strategy to adaptively balance different sources and alleviate negative transfer led by noisy data. Considering that data uncertainty could capture the noise inherent in the data, \ie, it reflects the reliability of output \cite{kendall2017what,chang2020data}, we could weigh different sources and samples based on the result of uncertainty estimation. As shown in Fig.~\ref{fig2}, $g_{\mu}(\cdot)$ and $g_{\sigma}(\cdot)$ serve on a classifier and uncertainty predictor, respectively. The output of network is modeled as a Gaussian distribution parameterized by mean $\mu$ and variance $\sigma$. Specifically, the mean is acted by original feature vector, while the variance quantifies uncertainty of training samples. For regression tasks, the Gaussian likelihood is defined as:
\begin{align}
	p(y_i|x_i)=\mathcal{N}(\mu_i,\sigma_i^2), 
\end{align}
with $\mu_i = h(x_i)=f(x_i)\circ g_{\mu}(x_i)$ and $\sigma_i = f(x_i)\circ g_{\sigma}(x_i)$. For classification task, we often squash the model output through a softmax function and obtain a scaled classification likelihood:
\begin{align}
	p(y_i|x_i,\sigma_i)=Softmax\bigl(\frac{1}{\sigma_i^2}h(x_i)\bigr).
\end{align}
This can be interpreted as a Boltzmann distribution (Gibbs distribution) and $\sigma_i^2$ works as temperature for re-scaling input. The log likelihood of output is: 
{\small \begin{align}
		log\,p(y_i=c|x_i,\sigma_i)=\frac{1}{\sigma_i^2}h_c(x_i)-log\sum_{c'\neq c}exp(\frac{1}{\sigma_i^2}h_{c'}(x_i)),
\end{align}}where $h_c(x_i)$ denotes the $c$-th element of vector $h(x_i)$. Then the total classification loss is defined as:
{\small \begin{equation}
		\label{eq13}
		\begin{split}
			\mathcal{L}_{cls}(\theta)&=-\frac{1}{M}\sum_{m=1}^{M}\frac{1}{N_m}\sum_{i=1}^{N_m}log\,p(y_i=c|x_i,\sigma_i)\\
			=\frac{1}{M}&\sum_{m=1}^{M}\frac{1}{N_m}\sum_{i=1}^{N_m}\frac{1}{\sigma_i^2}\mathcal{L}_{CE}(\theta)+log\frac{\sum_{c'}exp\bigl(\frac{1}{\sigma_i^2}h_{c'}(x_i)\bigr) }{\bigl(\sum_{c'}exp\bigl(h_{c'}(x_i)\bigr)  \bigr)^\frac{1}{\sigma_i^2}}\\
			\approx \frac{1}{M}&\sum_{m=1}^{M}\frac{1}{N_m}\sum_{i=1}^{N_m}\frac{1}{\sigma_i^2}\mathcal{L}_{CE}(\theta)+log\sigma_i,
		\end{split}
\end{equation}}where $\mathcal{L}_{CE}(\theta)$ denotes classification cross entropy loss with $h(x_i)$ not scaled. $M$ and $N_m$ denote the number of domains and samples, respectively. $log\sigma_i$ prevents $\sigma$ from getting too large. Noisy data with large uncertainty would be assigned less weights, \ie, ${1}/{\sigma^2}$. The derivation process of Eq.~\ref{eq13} is provided in supplementary material.\\

\subsection{Objective function}
\label{sec3.4}
The overall objective function of the proposed model is as follows:
\begin{equation}\label{eq10}
	\begin{split}
		\mathcal{L}_{total} =\mathcal{L}_{cls}({\theta})+ \lambda\left \| \mathcal{G} \right \|_\circledast,\\
		s.t.\;\mathcal{G}=\Psi_R(G^{\mathcal{S}_1}, \cdots, G^{\mathcal{S}_M}, G^{\mathcal{T}}),
	\end{split}
\end{equation}
where $\circledast$ denotes two operations: tensor rotation and tensor nuclear norm, and $\Psi_R$ represents the operation of stacking up all the domain-specific prototypical similarity matrices into a tensor. $\theta$ denotes neural network parameters.
The first term is classification loss and the second term imposes TLR constraint on the stacked tensor, aiming at achieving high-order alignment of domains. \par
{\tiny \begin{algorithm}[!t]
	\SetKwInOut{Input}{Input}\SetKwInOut{Output}{Output}
	\caption{Optimization of T-SVDNet}\label{alg2}
	\Input{Training data $\mathcal{S}_1,\cdots, \mathcal{S}_M, \mathcal{T}$ \;}
	\Output{Model parameters $\theta$ of T-SVDNet \;}
	\For{$iter=1$ to $max\_iter$}
	{	$\bullet$ \textbf{Updating} $\theta$\\
		{$\theta_f \leftarrow \theta_f - {\partial (\mathcal{L}_{cls}+\frac{\eta}{2}\left \| \mathcal{A}-\mathcal{G} \right \|_{F}^{2})}/{\partial \theta_f}; $ }\\
		$\theta_{\mu} \leftarrow \theta_{\mu} -{\partial \mathcal{L}_{cls}}/{\partial \theta_{\mu}};$ $\theta_{\sigma} \leftarrow \theta_{\sigma} -{\partial \mathcal{L}_{cls}}/{\partial \theta_{\sigma}};$\\
		$\bullet$ \textbf{Updating} $\mathcal{A}$\\
		$\mathcal{G}_{Rot}=Rotate(\mathcal{G});$ \\
		$\mathcal{G}_f = \text{FFT}(\mathcal{G}_{Rot},3)$\;
		\For{$k=1:n_3$}{$[\mathcal{U}_{f}^{(k)},\mathcal{S}_{f}^{(k)}, \mathcal{V}_{f}^{(k)}] = \text{SVD}(\mathcal{G}_{f}^{(k)})$\;
			$\mathcal{A}_f^{(k)}=\mathcal{U}_f^{(k)}\cdot D_{\frac{\lambda}{\eta}}( \mathcal{S}_f^{(k)})\cdot\mathcal{V}_f^{(k)T}$	\;}
		$\mathcal{A}_{Rot} = \text{IFFT}(\mathcal{A}_f, 3);$ \\
		$\mathcal{A} = Rotate(\mathcal{A}_{Rot})$ \;
		$\eta = min(\rho\eta, \eta_{max})$
	}
\end{algorithm}}
\subsection{Optimization of T-SVDNet}
\label{sec3.5}
The optimization of T-SVDNet is presented in Alg.~\ref{alg2}. In order to make the problem tractable, we introduce an auxiliary variable and alternatively update it along with the network parameters till convergence.\par
\textbf{Auxiliary variable.} To optimize the objective function in Eq.~\ref{eq10}, we first introduce an auxiliary tensor $\mathcal{A}$ to replace $\mathcal{G}$, which converts the original optimization problem into the following one:  
\begin{align}
	\underset{\theta, \mathcal{A}}{min}\mathcal{L}_{cls}+\lambda\left \| \mathcal{A} \right \|_\circledast+\frac{\eta}{2}\left \| \mathcal{A-G} \right \|_F^2,
\end{align}
where $\eta$ is a penalty parameter. It starts from a small initial positive scalar $\eta_0$, and gradually increases to the maximum truncated value $\eta_{max}$ with iterations, \ie, it is updated by $\eta = min(\rho\eta, \eta_{max})$, where $\rho$ represents the rate of increase which is set as $1.1$ in all experiments. The reason why we update $\eta$ in such a incremental fashion is that randomly initialized $\mathcal{A}$ may lead to the wrong direction of gradient descent at the beginning of training process.  \par 
\textbf{Update of network parameters $\theta$.} The parameters of feature extractor $\theta_f$, classifier $\theta_{\mu}$, and uncertainty predictor $\theta_{\sigma}$ are updated through gradient descent with auxiliary tensor $\mathcal{A}$ fixed. \par
\textbf{Update of auxiliary variable $\mathcal{A}$.} When network parameters are fixed, we optimize the subproblem associated with $\mathcal{A}$ as follows:
\begin{align}
	\underset{\mathcal{A}}{min}\lambda\left \| \mathcal{A} \right \|_\circledast+\frac{\eta}{2}\left \| \mathcal{A}-\mathcal{G} \right \|_{F}^{2}.
\end{align}  
We solve this problem in Fourier domain with basic procedure similar to Alg.~\ref{alg1}. We first transform tensor $\mathcal{G}_{Rot}$ to Fourier domain $\mathcal{G}_f$, and perform matrix SVD on each $k$-th frontal slice $\mathcal{G}_f^{(k)}$ and obtain the $\mathcal{U}_f^{(k)}$, $\mathcal{S}_f^{(k)}$, $\mathcal{V}_f^{(k)}$. Then, each frontal slice $\mathcal{A}_f^{(k)}$ of auxiliary variable can be updated by \textit{shrinkage operation} \cite{cai2010a,kolda2009tensor} on $\mathcal{G}_f^{(k)}$ in the Fourier domain defined as follows:  
\begin{table*}[h]
	\begin{center}
		\scalebox{0.7}{
			\begin{tabular}{lccccccc}
				\toprule[1.5pt]
				\multicolumn{1}{c}{Standards}   & \multicolumn{1}{c}{Methods}           & \multicolumn{1}{c}{$\to$ mm}         & \multicolumn{1}{c}{$\to$ mt}         & \multicolumn{1}{c}{$\to$ up}         & \multicolumn{1}{c}{$\to$ sv}         & \multicolumn{1}{c}{$\to$ syn}        & \multicolumn{1}{c}{Avg}        \\ \toprule[1.5pt]
				\multicolumn{1}{c}{\multirow{5}{*}{Single Best}} & Source-Only       &  $52.90\pm0.60 $          &  $97.20\pm0.60$          &  $84.70\pm0.80 $          &  $77.70\pm0.80$          &  $85.20\pm 0.60$          &  $80.80$          \\
				\multicolumn{1}{c}{}                             & DAN               & $63.78\pm 0.71$ & $96.31 \pm 0.54$ & $94.24\pm 0.87$ & $62.45\pm 0.72$ & $85.43\pm 0.77$ & $80.44$ \\
				\multicolumn{1}{c}{}                             & DANN              & $71.30\pm 0.56$ & $97.60\pm 0.75$ & $92.33\pm 0.85$ & $63.48\pm 0.79$ & $85.34\pm 0.84$ & $82.01$ \\
				\multicolumn{1}{c}{}                             & CORAL             & $62.53\pm 0.69$ & $97.21\pm 0.83$ & $93.45\pm 0.82$ & $64.40\pm 0.72$ & $82.77\pm 0.69$ & $80.07$ \\
				\multicolumn{1}{c}{}                             & ADDA              & $71.57\pm 0.52$ & $97.89\pm 0.84$ & $92.83\pm 0.74$ & $75.48\pm 0.48$ & $86.45\pm 0.62$ & $84.84$ \\ \hline
				\multirow{6}{*}{\begin{tabular}[c]{@{}c@{}}Source \\ Combination\end{tabular}}                  & Source-Only       & $63.37\pm 0.74$ & $90.50\pm 0.83$ & $88.71\pm 0.89$ & $63.54\pm 0.93$ & $82.44\pm 0.65$ & $77.71$      \\
				& DAN               & $67.91\pm 0.82$ & $97.52\pm 0.60$ & $93.45\pm 0.41$ & $67.79\pm 0.62$ & $86.90\pm 0.50$ & $82.69$      \\
				& DANN              & $70.80\pm 0.77$ & $97.91\pm 0.69$ & $93.53\pm 0.76$ & $68.54\pm 0.52$ & $87.40\pm 0.90$ & $83.61$      \\
				& ADDA              & $72.32\pm 0.74$ & $97.88\pm 0.60$ & $93.10\pm 0.79$ & $75.02\pm 0.80$ & $86.69\pm 0.55$ & $85.02$      \\
				& MCD               & $72.50\pm 0.67$ & $96.21\pm 0.81$ & $95.33\pm 0.74$ & $78.89\pm 0.78$ & $87.47\pm 0.65$ & $86.10$      \\
				& JAN               & $65.88\pm 0.68$ & $97.21\pm 0.73$ & $95.42\pm 0.77$ & $75.27\pm 0.71$ & $86.55\pm 0.64$ & $84.07$     \\ \hline 
				\multicolumn{1}{c}{\multirow{6}{*}{Multi-Source}}                    & MDAN              & $69.48\pm 0.30$ & $98.04 \pm 0.89$ & $92.39\pm 0.71$ & $69.23\pm 0.62$ & $87.44\pm 0.45$ & $83.30$      \\
				& MDDA              & {\color{green}$\bf 78.63\pm 0.61$} & {\color{green}$\bf 98.78\pm 0.42$} & $93.91\pm 0.48$ & $79.33\pm 0.80$ & {\color{green}$\bf 89.71\pm 0.72$} & {\color{green}$\bf88.12 $}     \\
				& DCTN              & $70.53\pm 1.24$ & $96.23\pm 0.82$ & $92.81\pm 0.27$ & $77.61\pm 0.41$ & $86.77\pm 0.78$ & $84.79$      \\
				& M$^3$SDA             & $72.82\pm 1.13$ & $98.43\pm 0.68$ & {\color{green}$\bf96.14\pm 0.81$} & {\color{green}$\bf81.32\pm 0.86$} & $89.58\pm 0.56$ & $87.65$      \\
				& T-SVDNet$_{part}$ & {\color{blue}$\bf{90.05\pm0.91}$}      & {\color{blue}$\bf99.24\pm0.08$}      & {\color{blue}$\bf98.61\pm0.16 $}      & {\color{blue}$\bf84.03\pm1.22$}      & {\color{blue}$\bf94.92\pm0.17$}      & {\color{blue}$\bf93.37$}      \\
				& T-SVDNet$_{all}$          & {\color{blue}$\bf91.22\pm0.74$}     & {\color{blue}$\bf99.28\pm0.11$}      & {\color{blue}$\bf98.63\pm0.22 $}            & {\color{blue}$\bf84.86\pm1.47$}      & {\color{blue}$\bf95.71\pm0.30$}      & {\color{blue}$\bf93.94$}           \\
				\bottomrule[1.5pt]
			\end{tabular}
		}	
	\end{center}
	\caption{{\small{Classification results on Digits-Five. The top value is highlighted in {\color{blue}{blue}} bold font and the second best in {\color{green} green} bold font.}}}
	\label{tab1}
\end{table*}
\begin{align}
	\mathcal{A}_f^{(k)}=\mathcal{U}_f^{(k)}\cdot D_{\frac{\lambda}{\eta}}( \mathcal{S}_f^{(k)})\cdot\mathcal{V}_f^{(k)T},
\end{align}
where {\small{${\small D_{{\lambda}/{\eta}}(\mathcal{S}_f^{(k)})=\mathcal{S}_f^{(k)}\cdot \mathcal{J}_f^{(k)}}$}} is singular value shrinkage operator. {\small{$\mathcal{J}_f^{(k)}$}} is a diagonal matrix with the $i$-th diagonal element to be {\small$\mathcal{J}_{f}^{(k)}(i,i)=(1-\frac{\lambda}{\eta\mathcal{S}_{f}^{(k)}(i,i)})_+$}. Finally, updated $\mathcal{A}$ is obtained by inverse fast Fourier transform from $\mathcal{A}_f$.

\section{Experiments}
In this section, we perform extensive evaluations on several benchmark datasets with state-of-the-art methods. 
\subsection{Datasets}
\textbf{Digits-Five}~\cite{hull1994database} contains 5 different domains including MNIST (mt), MNIST-M (mm), SVHN (sv), USPS (up), and Synthetic Digits (syn). Each domain consists of 10 numerals from `0' to `9'. Previous methods only use a subset of samples in each domain, \ie, 25000 training data and 9000 testing data. But we find that there will be further performance gain if all data are employed for training. For a fair comparison, we report the results on both settings (T-SVDNet$_{part}$ and T-SVDNet$_{all}$ in Tab.~\ref{tab1}).\par
\textbf{PACS}~\cite{li2017deeper} is a small-scale multi-domain dataset containing 9991 images from 4 domains: photo~(P), art-painting~(A), cartoon~(C), sketch~(S) whose styles are different. These domains share the same seven categories.\par
\textbf{DomainNet}~\cite{peng2019moment} is a large-scale dataset for Multi-Source Domain Adaptation. Due to the great number of categories and samples (345 categories, around 0.6 million images) and large domain shift. DomainNet is by far the most difficult dataset which contains 6 different domains: clipart (clp), infograph (inf), painting (pnt), quickdraw (qdr), real (rel), and sketch (skt).
\subsection{Compared methods}
For all experiments, we compare our method with state-of-the-art single-source and multi-source domain adaptation algorithms. Specifically, two strategies are adopted to train the single-source model: Single Best and Source Combination. The former reports the best result among all domains, while the latter simply combines all source domains together. Overall, these compared methods can be roughly categorized into two main groups: (1) adversarial-based methods include Domain Adversarial Neural Network (DANN) \cite{ganin2016domain}, Adversarial Discriminative Domain Adaptation (ADDA) \cite{tzeng2017adversarial}, Maximum Classifier Discrepancy (MCD) \cite{saito2018maximum}, Deep Cocktail Network (DCTN) \cite{xu2018deep}, Adversarial Multiple Source Domain Adaptation (MDAN) \cite{zhao2018adversarial} and Multi-Source Distilling Domain Adaptation (MDDA) \cite{zhao2020multi}; (2) another typical strategy is discrepancy minimization, the representative methods involve Deep Adaptation Network (DAN) \cite{long2015learning}, Joint Adaptation Network (JAN) \cite{long2017deep}, Residual Transfer Network (RTN) \cite{long2016unsupervised}, Correlation Alignment (CORAL) \cite{sun2017correlation}, and Moment Matching for Multi-Source Domain Adaptation (M$^3$SDA) \cite{peng2019moment}. Source-Only directly transfers the model trained in source domain to target domain. For a fair comparison, we use the same model architecture and data pre-processing routines as compared methods in all experiments. More implementation details are provided in supplementary materials.
\begin{table}[]
	\begin{center}
		\scalebox{0.7}{
			\begin{tabular}{cccccc}
				\toprule[1.5pt]
				\multicolumn{1}{c}{Methods} & \multicolumn{1}{l}{$\to$ A} & \multicolumn{1}{c}{$\to$ C} & \multicolumn{1}{c}{$\to$ S} & \multicolumn{1}{c}{$\to$ P} & \multicolumn{1}{c}{Avg} \\ \toprule[1.5pt]
				Source-Only  				& $75.97$					&$73.34$
				&$64.23$					&$91.65$					&$76.30$	\\
				MDAN                        & $83.54$                  & $82.34$                 & $72.42$                  & $92.91$                   & $82.80$ \\
				MDDA                        & {\color{green}$\bf86.73$ }                 & $86.24$                  & {\color{green}$\bf77.56$}                  & $93.89$  &                  {\color{green}$\bf86.11$}\\
				DCTN                        & {$84.67$}                  & {\color{green}$\bf 86.72$}                  & $71.84$                  & {\color{green}$\bf95.60$}        &$84.71$            \\
				M$^{3}$SDA                       & $84.20$                  & $85.68$                  & $74.62$                  & $94.47$                    &$84.74$ \\
				T-SVDNet                    & {\color{blue}$\bf90.43$}                  & {\color{blue}$\bf90.61$}                  & {\color{blue}$\bf85.49$}                  & {\color{blue}$\bf98.50$}                  &
				{\color{blue}$\bf91.25$} \\\bottomrule[1.5pt]
\vspace{-1em}		\end{tabular}}
	\end{center}
	\caption{{\small{Classification results on PACS. The top value is highlighted in {\color{blue}{blue}} bold font and the second best in {\color{green} green} bold font.}}}
	\label{tab2}
\end{table}
\begin{table*}[]
	\begin{center}
		\scalebox{0.7}{
			\begin{tabular}{ccccccccc}
				\toprule[1.5pt]
				Standards                       & \multicolumn{1}{c}{Methods} & $\to$clp      & $\to$inf      & $\to$pnt      & $\to$qdr      & $\to$rel      & $\to$skt      & Avg  \\ \toprule[1.5pt]
				\multirow{7}{*}{\begin{tabular}[c]{@{}c@{}}Single- \\ Best\end{tabular}}    & Source-Only                 & $39.6\pm 0.6$ & $8.2\pm 0.8$  & $33.9\pm 0.6$ & $11.8\pm 0.7$ & $41.6\pm 0.8$ & $23.1\pm0.7$     & $26.4$ \\
				& DAN                         & $39.1\pm 0.5$ & $11.4\pm 0.8$ & $33.3\pm 0.6$ & {\color{green}$\bf16.2\pm 0.4$} & $42.1\pm 0.7$ & $29.7\pm 0.9$ & $28.6$ \\
				& RTN                         & $35.3\pm 0.7$ & $10.7\pm 0.6$ & $31.7\pm 0.8$ & $13.1\pm 0.7$ & $40.6\pm 0.6$ & $26.5\pm 0.8$ & $26.3$ \\
				& JAN                         & $35.3\pm 0.7$ & $9.1\pm 0.6$  & $32.5\pm 0.7$ & $14.3\pm 0.6$ & $43.1\pm 0.8$ & $25.7\pm 0.6$ & $26.7$ \\
				& ADDA                        & $39.5\pm 0.8$ & $14.5\pm 0.7$ & $29.1\pm 0.8$ & $14.9\pm 0.5$ & $41.9\pm 0.8$ & $30.7\pm 0.7$ & $28.4$ \\
				& DANN                        & $37.9\pm 0.7$ & $11.4\pm 0.9$ & $33.9\pm 0.6$ & $13.7\pm 0.6$ & $41.5\pm 0.7$ & $28.6\pm 0.6$ & $27.8$ \\
				& MCD                         & $42.6\pm 0.3$ & $19.6\pm 0.8$ & $42.6\pm 1.0$ & $3.8\pm 0.6$  & $50.5\pm 0.4$ & $33.8\pm 0.9$ & $32.2$ \\ \hline
				\multirow{6}{*}{\begin{tabular}[c]{@{}c@{}}Source \\ Combination\end{tabular}} & Source-Only                 & $47.6\pm 0.5$ & $13.0\pm 0.4$ & $38.1\pm 0.5$ & $13.3\pm 0.4$ & $51.9\pm 0.9$ & $33.7\pm 0.5$ & $32.9$ \\
				& DAN                         & $45.4\pm 0.5$ & $12.8\pm 0.9$ & $36.2\pm 0.6$ & $15.3\pm 0.4$ & $48.6\pm 0.7$ & $34.0\pm 0.5$ & $32.1$ \\
				& RTN                         & $44.2\pm 0.6$ & $12.6\pm 0.7$ & $35.3\pm 0.6$ & $14.6\pm 0.8$ & $48.4\pm 0.7$ & $31.7\pm 0.7$ & $31.1$ \\
				& ADDA                        & $47.5\pm 0.8$ & $11.4\pm 0.7$ & $36.7\pm 0.5$ & $14.7\pm 0.5$ & $49.1\pm 0.8$ & $33.5\pm 0.5$ & $32.2$ \\
				& JAN                         & $40.9\pm 0.4$ & $11.1\pm 0.6$ & $35.4\pm 0.5$ & $12.1\pm 0.7$ & $45.8\pm 0.6$ & $32.3\pm 0.6$ & $29.6$ \\
				& MCD                         & $54.3\pm 0.6$ & $22.1\pm 0.7$ & $45.7\pm 0.6$ & $7.6\pm 0.5$  & $58.4\pm 0.7$ & $43.5\pm 0.6$ & $38.5$ \\ \hline
				\multirow{6}{*}{\begin{tabular}[c]{@{}c@{}}Multi- \\ Source\end{tabular}}   & MDAN                        & $52.4\pm 0.6$ & $21.3\pm 0.8$ & $46.9\pm 0.4$ & $8.6\pm 0.6$  & $54.9\pm 0.6$ & $46.5\pm 0.7$ & $38.4$ \\
				& MDDA                        & {\color{green}$\bf59.4\pm 0.6$} & $23.8\pm 0.8$ & {\color{green}$\bf53.2\pm 0.6$} & $12.5\pm 0.6$ & $61.8\pm 0.5$ & $48.6\pm 0.8$ & {\color{green}$\bf43.2$} \\
				& DCTN                        & $48.6\pm 0.7$ & $23.5\pm 0.6$ & $48.8\pm 0.6$ & $7.2\pm 0.5$  & $53.5\pm 0.6$ & $47.3\pm 0.5$ & $38.2$ \\
				& M$^3$SDA                       & $58.6\pm 0.5$ & {\color{blue}$\bf26.0\pm 0.9$} & $52.3\pm 0.6$ & $6.3\pm 0.6$  & {\color{green}$\bf62.7\pm 0.5$} & {\color{green}$\bf49.5\pm 0.8$} & $42.6$ \\
				& T-SVDNet                    & {\color{blue}$\bf66.1\pm 0.4$} & {\color{green}$\bf25.0\pm 0.8$} & {\color{blue}$\bf54.3\pm 0.7$} & {\color{blue}$\bf16.5\pm 0.9$} & {\color{blue}$\bf65.4\pm 0.5$} & {\color{blue}$\bf54.6\pm 0.6$} & {\color{blue}$\bf47.0$} \\ \bottomrule[1.5pt]
		\end{tabular}}
	\end{center}
	\caption{\small{Classification results on DomainNet. The top value is highlighted in {\color{blue}{blue}} bold font and the second best in {\color{green}green} bold font.}}
	\label{tab3}
\end{table*}

\begin{table*}[!htbp]
	\centering                                                                      \scalebox{0.7}{
		\begin{tabular}{lccccccc}
			\toprule[1.5pt]
			\multicolumn{1}{c}{Methods}            & \multicolumn{1}{c}{$\to$ mm}         & \multicolumn{1}{c}{$\to$ mt}         & \multicolumn{1}{c}{$\to$ up}         & \multicolumn{1}{c}{$\to$ sv}         & \multicolumn{1}{c}{$\to$ syn}        & \multicolumn{1}{c}{Avg}    & \multicolumn{1}{c}{Gain}    \\ \toprule[1.5pt]
			\multicolumn{1}{c}{Source-Only}       &  $67.25\pm0.81 $          &  $98.88\pm0.49$          &  $97.87\pm0.43 $          &  $77.76\pm0.92$          &  $92.41\pm 0.57$          &  $86.83$       &  $-$   \\
			\multicolumn{1}{c}{T-SVDNet~(+E)}       & $73.85\pm 0.84$ & $98.96\pm 0.35$ & $97.87\pm 0.65$ & $77.86\pm 0.84$ & $92.44\pm 0.39$ & $88.19$ & $\bf1.36\uparrow$\\
			\multicolumn{1}{c}{T-SVDNet~(+E+T)}                       & $88.76\pm 0.41$ & $99.16\pm 0.26$ & $98.09\pm 0.14$ & $82.94\pm 0.90$ & $94.47\pm 0.62$ & $92.68$ & $\bf5.85\uparrow$ \\
			\multicolumn{1}{c}{T-SVDNet~(+E+T+U)}                       &  {$91.22\pm0.74$}     & {$99.28\pm0.11$}      & {$98.63\pm0.22 $}            & {$84.86\pm1.47$}      & {$95.71\pm0.30$}      & {$93.94$}  & $\bf7.11\uparrow$ 
			\\ \bottomrule[1.5pt]
	\end{tabular}}
	\caption{{\small{Ablation study on key components of model on Digits-Five.}}} 
	\label{tab4}
	\vspace{-10pt}
\end{table*}
\subsection{Experimental results}
The results on \textbf{Digits-Five} are shown in Tab.~\ref{tab1}. Overall, our method tops the list in all domains and achieves 93.37\% average accuracy, around 5.25\% higher than the second best method MDDA. In particular, a performance improvement about 11.42\% and 5.21\% over MDDA is achieved on `$\to mm$' and `$\to syn$' tasks, respectively. The performance will be further boosted to 93.94\% if all training data is used, outperforming other algorithms by a large margin.  \par
The results on \textbf{PACS} are shown in Tab.~\ref{tab2}. Our method T-SVDNet achieves the best performance on all domains and gets 91.25\% average accuracy, outperforming the second best method MDDA by 5.14\%. Especially on `$\to S$' task, our method achieves a 7.93\% performance gain over MDDA. \par
The experimental results on \textbf{DomainNet} are reported in Tab.~\ref{tab3}. Overall, T-SVDNet achieves the best performance on five out of six tasks. It obtains average accuracy of 47.0\% on six domains and ranks the first in the list, with 3.8\% performance improvement over MDDA, which is mainly attributed to the thorough exploration of high-order relations between different domains and categories. It is noteworthy that the performances of many MDA methods drop obviously compared to Single Best on `$\to qdr$' task due to negative transfer, while our method still attains better performance because of uncertainty-aware weighting strategy. Negative transfer is avoided by filtering out noisy source samples near decision boundaries for training, while clean data with low noise intensity are fully exploited. \par


\begin{figure}
	\centering
	\includegraphics[scale=0.21]{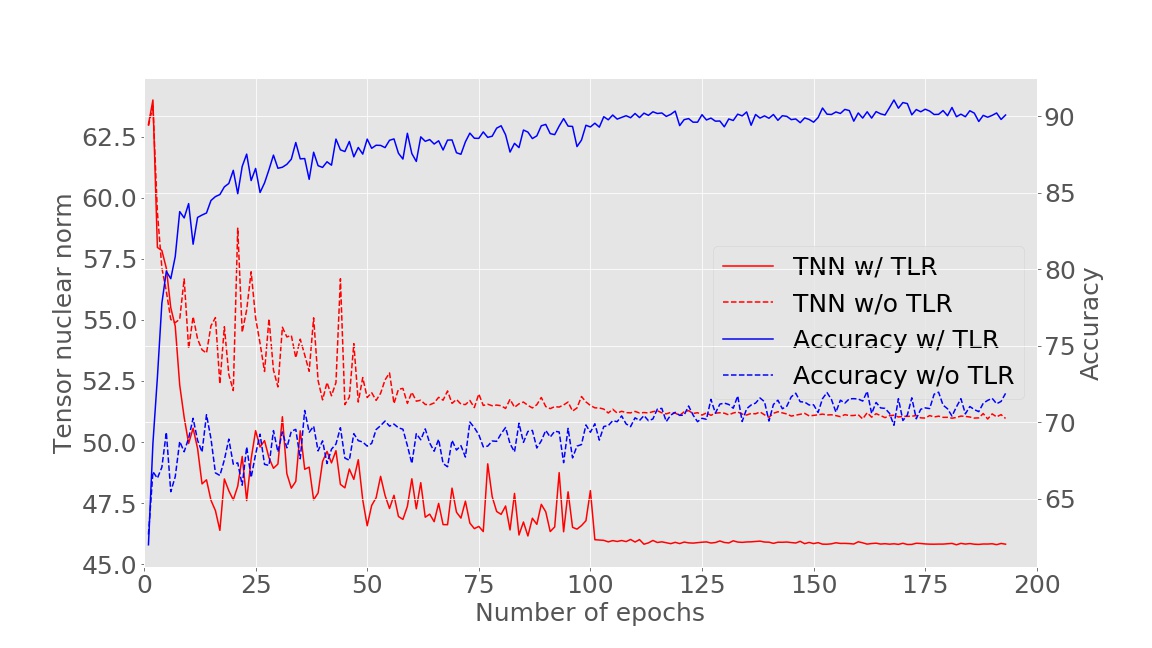}\\
	\caption{Tensor nuclear norm and classification accuracy curves on ``$\to$ mm'' task. }
	\label{fig6}
	\vspace{-1em}
\end{figure}
\section{Analysis}
\textbf{Ablation study.} We further validate the effects of some key components in our framework. Tab.~\ref{tab4} shows the results of controlled experiments on Digits-Five dataset. As a reference, we report the performance of Source-Only that directly transfers the model trained on source domains to target domain. For convenience, `+E', `+T', `+U' denote entropy minimization constraint on target domain, tensor-low-rank constraint, and uncertainty-aware weighting, respectively. In general, we have the following observations according to Tab.~\ref{tab4}: (1) Entropy minimization boosts performance obviously due to the exploitation of target domain; (2) It is remarkable that Tensor-Low-Rank constraint significantly improves the performance by 14.91\% on `$\to$ mm' task. This is attributed to the high-order alignment between different domains and the extraction of domain-invariant features; (3) Uncertainty-aware weighting strategy further improves the performance by 1.26\% on average, which suggests that our model is able to learn more transferable features across domains.\par

\textbf{The effect of TLR constraint.} We compute tensor nuclear norm (TNN) which is usually used as an approximate measure of tensor rank. As shown in Fig.~\ref{fig6}, we compare the TNN curves w/ and w/o TLR constraint. We find that TNN w/ TLR drops significantly during the first several epochs and stabilizes at around 46, while the baseline w/o TLR drops slowly and becomes stable earlier. This demonstrates that our proposed TLR constraint is effective and brings large performance improvement. 
\begin{figure}
	\centering
	\includegraphics[scale=0.15]{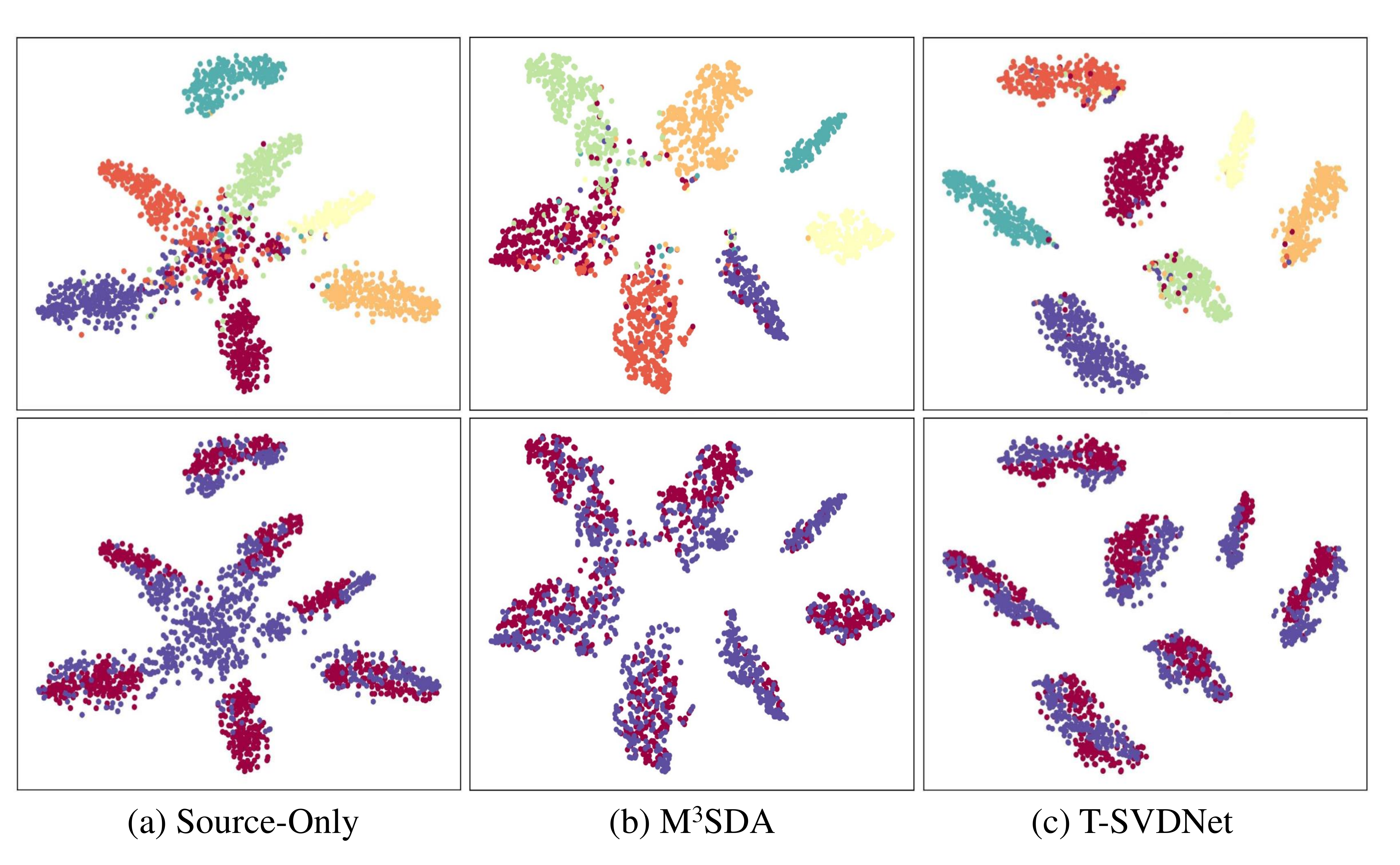}\\
	\caption{The t-SNE visualizations of feature embeddings on `$\to$C' task on PACS. The top row represents category information (each color denotes a class). The bottom row represents domain information (red: source domain; purple: target domain).}
	\label{fig3}
	\vspace{-1em}
\end{figure}

\begin{figure}
	\centering
	\includegraphics[scale=0.31]{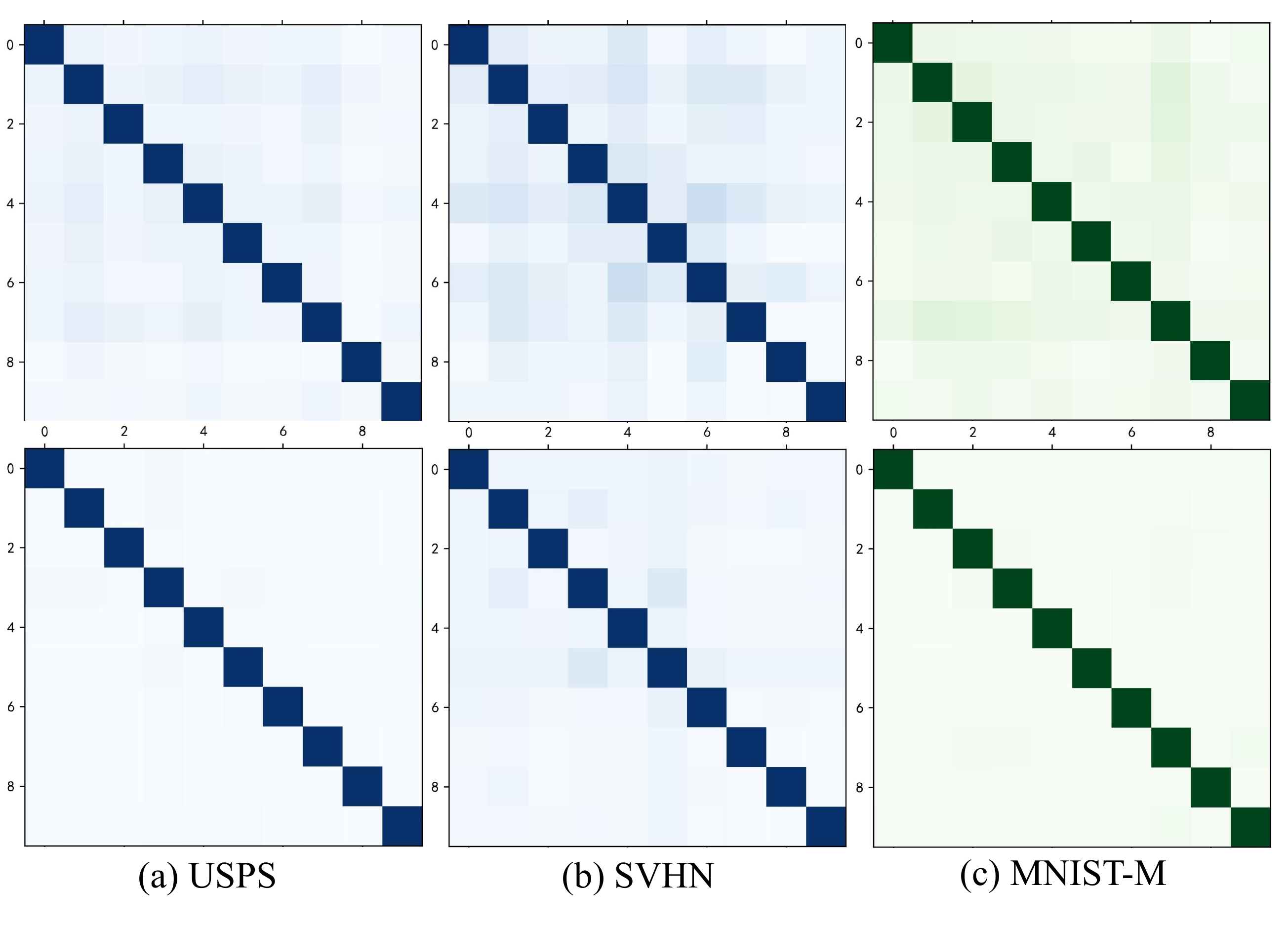}\\
	\caption{Visualizations of similarity matrices on `$\to$mm' task on Digits-Five. The top row denotes the model w/o TLR constraint, and the bottom row is T-SVDNet. Blue and green represent source and target domain, respectively.}
	\label{fig4}
	\vspace{-1em}
\end{figure}
\textbf{Feature visualization.} 
To demonstrate the transfer ability of our model, we visualize the feature embeddings of different models on `$\to C$' task on PACS. As shown in Fig.~\ref{fig3}~(a), the target features learned by Source-Only almost mismatch with source domain and different classes in target domain are entirely mixed up. Compared to M3SDA and Source-Only, our method produces clusters with clearer boundaries, which suggests that T-SVDNet possesses better transfer ability on target and is able to eliminate domain discrepancy without sacrificing discrimination ability.\par
\textbf{Visualizations of similarity matrices.} 
To further validate the effect of TLR constraint, we visualize the prototypical similarity matrices of three domains on Digits-Five dataset in Fig. \ref{fig4}. Compared to the baseline without TLR constraint (the top row), our method (the bottom row) could capture clearer category-wise data structure. Specifically, matrices in the bottom row contain less domain-specific noise, because we search for a lowest-rank structure of tensor and enforce prototypical correlations to be consistent across domains. Especially on the target domain (MNIST-M), the noise is reduced by a large margin compared to Source-Only. These results indicate the effectiveness of TLR constraint on aligning source and target domains.\par
\textbf{Uncertainty estimation.} We conduct qualitative and quantitative experiments to demonstrate the ability of model to measure noise intensity (data uncertainty).\par
\begin{figure}
	\centering
	\includegraphics[scale=0.3]{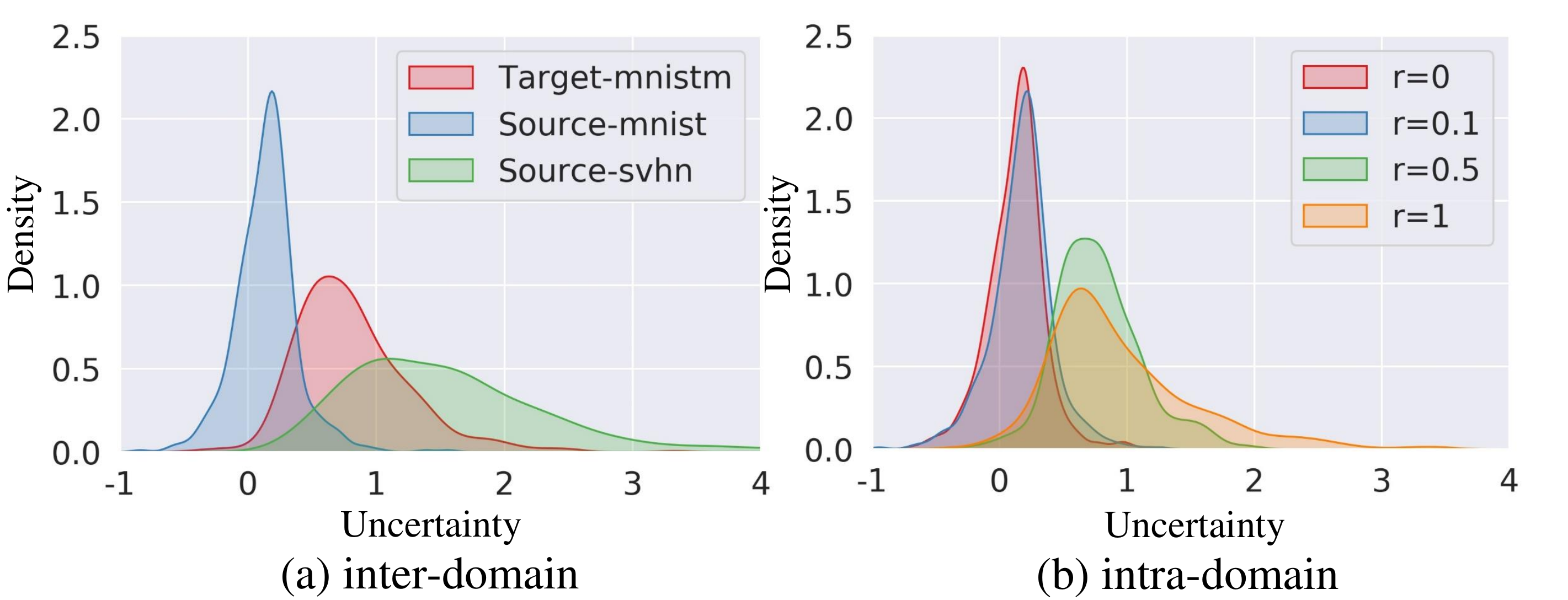}\\
	\caption{(a)~Uncertainty distributions of different domains on Digits-Five. (b)~Uncertainty distribution of single domain varies with the increasing noise intensity $r$ on MNIST.}
	\label{fig5}
	\vspace{-1em}
\end{figure}

\textit{(1)~Inter-domain weighting.}  The uncertainty distributions of different domains on `$\to$mm' task are shown in Fig.~\ref{fig5}~(a). Overall, the estimated uncertainty is highly correlated with domain quality. \eg, the uncertainty distribution of high-quality domain MNIST (blue curve) is more concentrated than low-quality domain SVHN (green curve). This validates that our model is able to measure the quality of domain, and guide the combination of data distributions.\par
\textit{(2)~Intra-domain weighting.} To demonstrate the ability of our model to capture noise inherent in data, we add different proportions of Gaussian noise to images and plot the estimated uncertainty distributions in Fig.~\ref{fig5}~(b). Specifically, we add noise sampled from Gaussian distribution $\mathcal{N}(0,\bf{I})$ to original images, \ie, $\tilde{x}_i = x_i + r\epsilon_i$, where $\epsilon$ denotes noise and $r$ controls the intensity of noise. According to Fig.~\ref{fig5}~(b), when noise intensity is small ($r=0.1$), the curves of noisy and clean samples ($r=0$) are highly overlapped. However, with the increase of noise intensity ($r=0.5,1$), the uncertainty distributions get more dispersed. This demonstrates that our model could accurately evaluate intra-domain data quality, so that noisy samples would be assigned less weights and negative transfer would be avoided.
\section{Conclusion}
In this paper, we propose the T-SVDNet for multi-source domain adaptation, which is featured by incorporating tensor singular value decomposition into neural network training process. Category-wise relations are modeled by prototypical similarity matrix, aiming at capturing complex data structure. Furthermore, high-order relations between different domains are fully explored by imposing tensor-low-rank constraint on the tensor stacked by domain-specific similarity matrices. In addition, a novel uncertainty-aware weighting strategy is proposed to combine data distributions of different domains, which reduces negative transfer led by noisy data. We adopt alternative optimization algorithm to train T-SVDNet efficiently. Extensive experiments on three public benchmark datasets demonstrate the favorable performance against state-of-the-art methods. 

{\small
	\bibliographystyle{ieee_fullname}
	\bibliography{egbib}
}
\section{Supplementary}
\subsection{Relevant Definitions}
{\definition (Tensor product) The tensor product $\mathcal{C}$ of $\mathcal{A}\in \mathbb{R}^{n_1\times n_2\times n_3}$ and $\mathcal{B}\in \mathbb{R}^{n_2\times n_4\times n_3}$, \ie, $\mathcal{C}=\mathcal{A}\mathcal{B}$, is a tensor of size $n_1\times n_4\times n_3$, each $(i,j)$-th tube of which denoted by $\mathcal{C}(i,j,:)$ with $i=1,2,\cdots,n_1$ and $j=1,2,\cdots,n_4$ is given by:}
\begin{align}
	\sum_{l}^{n_2}\mathcal{A}(i,l,:)\circ \mathcal{B}(l,j,:),
\end{align}
where $\circ$ denotes the circular convolution between two vectors. Tensor product in the original domain can be replaced by matrix multiplication of frontal slices in the Fourier domains as follows:
\begin{align}
	\mathcal{C}_f^{(k)} = \mathcal{A}_f^{(k)}\cdot\mathcal{B}_f^{(k)}.
\end{align}
{\definition (Tensor transpose) For $\mathcal{A}\in \mathbb{R}^{n_1\times n_2\times n_3}$, the transpose of $\mathcal{A}$ denoted by $\mathcal{A}^T\in \mathbb{R}^{n_2\times n_1\times n_3}$ can be obtained by transposing each frontal slice of $\mathcal{A}$ and reversing the order of the transposed slices along the third dimension.}
{\definition (Identity tensor) The identity tensor $\mathcal{I}\in \mathbb{R}^{n_1\times n_1\times n_3}$ is a tensor whose first frontal slice is the $n_1 \times n_1$ identity matrix and all other frontal slices are zeros.}
{\definition (Orthogonal tensor) A tensor $\mathcal{Q} \in \mathbb{R}^{n_1\times n_1\times n_3}$ is orthogonal if 
\begin{align}
	\mathcal{Q}^T*\mathcal{Q}=\mathcal{Q}*\mathcal{Q}^T=I,
\end{align}
where $*$ is the tensor product.}
\begin{table*}[t]
	\begin{tabular}{clccccccccc}
		\hline
		\multicolumn{2}{c}{dataset}                    & domains            & classes              & input size               & backbone                    & batch size          & learning rate              & $\lambda$                & $\mu_{max}$                   & feature dimension     \\ \hline
		\multicolumn{2}{c}{Digits-Five}                & 5                  & 10                   & 32*32                    & 3Conv-2FC                   & 128                 & 5e-4                       & 1000                  & 1                    & 2048                  \\
		\multicolumn{2}{c}{\multirow{2}{*}{PACS}}      & \multirow{2}{*}{4} & \multirow{2}{*}{7}   & \multirow{2}{*}{224*224} & \multirow{2}{*}{ResNet-18}  & \multirow{2}{*}{16} & E:3e-5                     & \multirow{2}{*}{1000} & \multirow{2}{*}{0.1} & \multirow{2}{*}{512}  \\
		\multicolumn{2}{c}{}                           &                    &                      &                          &                             &                     & C:1e-3                     &                       &                      &                       \\
		\multicolumn{2}{c}{\multirow{2}{*}{DomainNet}} & \multirow{2}{*}{6} & \multirow{2}{*}{345} & \multirow{2}{*}{224*224} & \multirow{2}{*}{ResNet-101} & \multirow{2}{*}{16} & E:5e-5                     & \multirow{2}{*}{100}  & \multirow{2}{*}{1}   & \multirow{2}{*}{2048} \\
		\multicolumn{2}{c}{}                           &                    &                      &                          &                             &                     & {C:5e-4} &                       &                      &                       \\ \hline	
	\end{tabular}
	\caption{The experimental setups on different datasets. E and C denote feature extractor and classifier, respectively.}
	\label{tab5}
\end{table*}
{\definition (f-diagonal tensor) The f-diagonal tensor is a tensor each frontal slice of which is diagonal matrix. The tensor product of two f-diagonal tensors with the same size $n_1\times n_2 \times n_3$ is also a tensor with the same size, each $i$-th $(i=1,\cdots,min(n_1, n_2))$ diagonal tube of which is:
\begin{align}
	\mathcal{C}(i,i,:)=\mathcal{A}(i,i,:)\circ \mathcal{B}(i,i,:).
\end{align}
}
\subsection{Derivation of Eq.~\ref{eq13}}
First, the cross entropy loss $\mathcal{L}_{CE}$ is denoted as:
\begin{equation}
	\begin{split}
		\mathcal{L}_{C E} &=-\log \left(\operatorname{Softmax}\left(h\left(x_{i}\right)\right)\right) \\
		&=-h_{c}\left(x_{i}\right)+\log \sum_{c^{\prime}} \exp \left(h_{c^{\prime}}\left(x_{i}\right)\right),
\end{split}
\end{equation}
then we obtain the following derivation process pf Eq.~\ref{eq13}:
\begin{equation}
	\begin{split}
	&-\log p\left(y_{i}=c \mid x_{i}, \sigma_{i}\right) \\
	=&-\frac{1}{\sigma_{i}^{2}} h_{c}\left(x_{i}\right)+\frac{1}{\sigma_{i}^{2}} \log \sum_{c^{\prime}} \exp \left(h_{c^{\prime}}\left(x_{i}\right)\right) \\
	&+\log \sum_{c^{\prime}} \exp \left(\frac{1}{\sigma_{i}^{2}} h_{c^{\prime}}\left(x_{i}\right)\right)-\frac{1}{\sigma_{i}^{2}} \log \left(\sum_{c^{\prime}} \exp \left(h_{c^{\prime}}\left(x_{i}\right)\right)\right) \\
	=& \frac{1}{\sigma_{i}^{2}} \mathcal{L}_{C E}(\theta)+\log \frac{\sum_{c^{\prime}} \exp \left(\frac{1}{\sigma_{i}^{2}} h_{c^{\prime}}\left(x_{i}\right)\right)}{\left(\sum_{c^{\prime}} \exp \left(h_{c^{\prime}}\left(x_{i}\right)\right)\right)^{\frac{1}{\sigma_{i}^{2}}}} \\
	\approx & \frac{1}{\sigma_{i}^{2}} \mathcal{L}_{C E}(\theta)+\log \sigma_{i},
\end{split}
\end{equation}
here we introduce a simplifying assumption in the last transition:
{\small \begin{align}
\sum_{c^{\prime}} \exp \left(\frac{1}{\sigma_{i}^{2}} h_{c^{\prime}}\left(x_{i}\right)\right) \approx \sigma_{i}\left(\sum_{c^{\prime}} \exp \left(h_{c^{\prime}}\left(x_{i}\right)\right)\right)^{\frac{1}{\sigma_{i}^{2}}}
\end{align}}
which becomes equality when $\sigma_i\to 1$. This assumption simplifies the optimization objective, while the performance is improved empirically. 

\subsection{Implementation Details}
Overall, for fair comparisons, we use the same model architecture and data pre-processing routines as compared methods in all experiments. Specifically, we present the detailed parameter settings on three datasets in Tab.~\ref{tab5}. Parameters $\gamma$ and $\rho$ are set as $0.05$ and $1.1$ in all experiments.

\end{document}